\pdfoutput=1
\documentclass[11pt]{article}

\usepackage{EMNLP2023}

\usepackage{amsmath}
\usepackage{multirow}
\usepackage{graphicx}
\usepackage{colortbl}
\usepackage{arydshln} 
\usepackage{color}
\usepackage{makecell}
\usepackage{amsfonts,amssymb}
\usepackage[ruled, vlined, linesnumbered]{algorithm2e}
\usepackage{subfigure}
\usepackage{booktabs}
\usepackage{makecell}

\usepackage{float}
\usepackage{amsfonts}
\usepackage{enumitem}
\usepackage{tikz}
\usepackage{tcolorbox}
\usepackage{siunitx}
\usepackage{amssymb}%
\usepackage{pifont}%

\usepackage{times}
\usepackage{latexsym}
\usepackage[figuresright]{rotating}

\usepackage[T1]{fontenc}
\usepackage[utf8]{inputenc}
\usepackage{cleveref}
\crefname{section}{§}{§§}
\Crefname{section}{§}{§§}

\usepackage{microtype}

\usepackage{inconsolata}

\NewDocumentCommand{\heng}
{ mO{} }{\textcolor{red}{\textsuperscript{\textit{Heng}}\textsf{\textbf{\small[#1]}}}}

\NewDocumentCommand{\yang}
{ mO{} }{\textcolor{blue}{\textsuperscript{\textit{Yang}}\textsf{\textbf{\small[#1]}}}}

\title{The Shifted and The Overlooked:\\A Task-oriented Investigation of User-GPT Interactions}

\author{Siru Ouyang\textsuperscript{1\thanks{\ \ Work partially done during internship at Microsoft.}}, Shuohang Wang\textsuperscript{2}, Yang Liu\textsuperscript{2}, Ming Zhong\textsuperscript{1}, Yizhu Jiao\textsuperscript{1}, Dan Iter\textsuperscript{2} \\
\textbf{Reid Pryzant\textsuperscript{2}}, \textbf{Chenguang Zhu\textsuperscript{2}}, \textbf{Heng Ji\textsuperscript{1}}, \textbf{Jiawei Han\textsuperscript{1}} \\
  \textsuperscript{1} University of Illinois Urbana-Champaign 
  \textsuperscript{2} Microsoft Azure AI \\
  \normalsize\texttt{siruo2@illinois.edu} \\}

\begin{document}
\maketitle
\begin{abstract}
Recent progress in Large Language Models (LLMs) has produced models that exhibit remarkable performance across a variety of NLP tasks. However, it remains unclear whether the existing focus of NLP research accurately captures the genuine requirements of human users. This paper provides a comprehensive analysis of the divergence between current NLP research and the needs of real-world NLP applications via a large-scale collection of user-GPT conversations. 
% \daniter{ShareGPT is a biased sample of "real-world AI apps" so this might be an overstatement.}
% We analyze real user queries to GPT, comparing these data against NLP benchmark tasks and revealing a large number of tasks
% which are over-represented in user interactions but critically overlooked in traditional academic research benchmarks, e.g., emotional support and meal planning.
We analyze a large-scale collection of real user queries to GPT.
We compare these queries against existing NLP benchmark tasks and identify a significant gap between the tasks that users frequently request from LLMs and the tasks that are commonly studied in academic research. 
For example, we find that tasks such as ``design'' and ``planning'' are  prevalent in user interactions but are largely neglected or different from traditional NLP benchmarks.
We investigate these overlooked tasks, dissect the practical challenges they pose, and provide insights toward a roadmap to make LLMs better aligned with user needs.
% We also propose potential solutions alongside a roadmap for directing NLP research toward the most pressing and neglected practical challenges posed by state-of-the-art LLMs.
\end{abstract}

\section{Introduction}

% \heng{what should be done? stake holders vs. regular users. }

% \reid{rewrite of the version below , i think this is a good story but not sure whether it's entirely true, people still work on e.g. sentiment analysis now, so need to be careful about the wording and/or include citations}
Over the past years, the NLP community has witnessed several paradigm shifts in technology followed by renewed research focus on applications that test the limits of this technology \cite{sun2022paradigm}. For example, distributed word representations \cite{landauer1998introduction, mikolov2013efficient} enabled a better characterization of the semantic similarity between words, entailing NLP research gravitated towards tasks like sentiment analysis and dependency parsing \cite{klein2003accurate}. Subsequent technologies like the transformer architecture \cite{vaswani2017attention} and contextual word representations \cite{devlin2018bert,peters-etal-2018-deep} further expanded the space of possible applications and the edge of NLP research, such as machine translation \cite{bahdanau2014neural} and document summarization \cite{tan2017abstractive}.

Most recently, large language models (LLMs) \cite{brown2020language, chowdhery2022palm} such as ChatGPT, emerged as powerful tools capable of achieving unprecedented success across a broad spectrum of NLP tasks~\cite{hendryckstest2021, clark2018think}. These models have become accessible and popular among non-NLP experts, opening the door for many new user applications.

The flood of new applications and the sharing of user interactions with LLMs~\cite{Tay} provide a great opportunity to closely examine the distribution of real applications users need on a daily basis. 
After a detailed analysis, we identify a conspicuous gap between real-world user queries and established NLP benchmarks, suggesting another shift in NLP focus is needed. To systematically analyze the phenomenon and to bridge the gap, we conduct a battery of experiments aiming to examine the following aspects:

\begin{itemize}
    \item What is the distribution of real-world user queries in terms of domain and task types, and how do they shift from traditional NLP benchmarks (\cref{sec:RQ1})?

    \item What are the emerging tasks and requirements from real-world user queries that may be overlooked in previous studies (\cref{sec:RQ2})?

\end{itemize}

% \daniter{I think the above list would be better used to describe the findings rather than to list the questions. I see that it is a summary of the sections, but the section headers should speak for themselves and instead, we should use the intro to succinctly present our findings.}

% \heng{I think the point is not the previous tasks are not reflecting 'real-world' problems; they are but they target for specialized users who are performing certain tasks, for example for analysts to perform intelligence analysis, or for biomedical scientists to understand literature, etc. I think you are advocating for targeting a much wider range of users and use NLP to assist their daily tasks, e.g., help farmers to control weeds and insects?}

We start by investigating ShareGPT\footnote{\url{https://sharegpt.com/}}, a large-scale collection of user-GPT conversations in the real world, containing 94,145 split data samples. ShareGPT has been used for training powerful LLMs~\cite{vicuna2023, xu2023wizardlm} and incorporated into new datasets~\cite{zheng2023lmsys, gudibande2023false}, both showing substantial advantages. Specifically, we design an annotation framework where we employ GPT-4~\cite{OpenAI2023GPT4TR} to generate the related information for every user query that appears in ShareGPT. We subsequently delve into the obtained data and conduct comprehensive analyses to answer the aforementioned questions\footnote{Code is available at \url{https://github.com/ozyyshr/ShareGPT_investigation}.}. We summarize our key findings as follows:

% \noindent
% 1) Generally, real-world user queries demonstrate a tendency towards more aligned with daily life with enlarging diverse user bases.

% \noindent
% 2) We discovered several tasks, including providing advice, designing, planning, etc., that are seldom touched and pose new requirements in the era of LLM. 

% \noindent
% 3) We summarized the shifting trends and challenges, providing insights to fill the gap for both stakeholders and users.
\begin{enumerate}
    \item Generally, real-world user queries demonstrate a tendency towards more aligned with daily life with enlarging diverse user bases.
    \vspace{-3mm}
    \item We discovered several tasks, including providing advice, designing, planning, etc., that are seldom touched and pose new requirements in the era of LLM. 
    \vspace{-3mm}
    \item We summarized the shifting trends and challenges, providing insights to fill the gap for both stakeholders and users.
\end{enumerate}

\section{Methodology}

\begin{figure*}[hbt]
\centering
\includegraphics[width=1\textwidth]{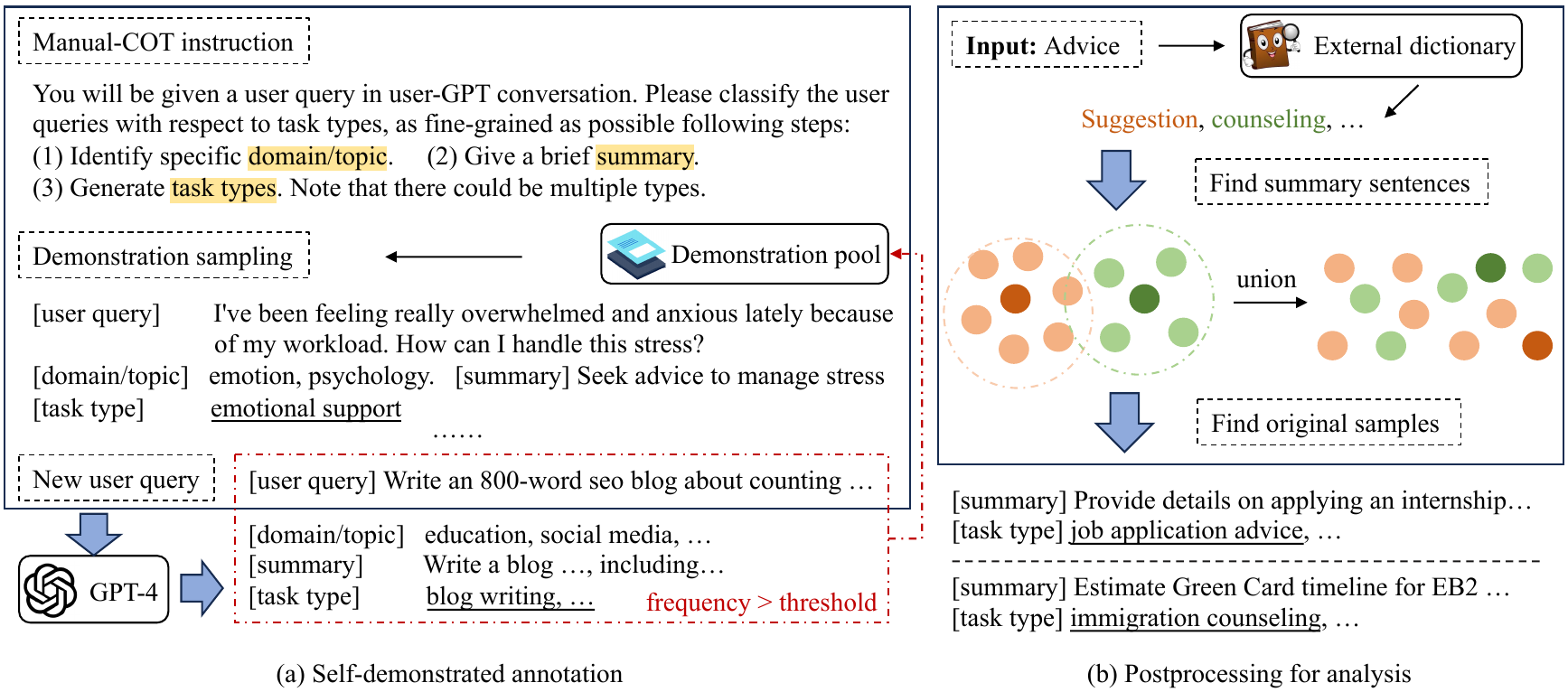}
\caption{ General framework of how we conduct step-to-step and self-demonstrated prompting for annotation using GPT-4 (a), and the post-processing for certain words/phrases (b).}
\label{framework}
\vspace*{-3mm}
\end{figure*}

In this section, we employ GPT to annotate the topic/domain and task type of every sample in the ShareGPT collection. The whole annotation and post-processing pipeline is illustrated in Figure~\ref{framework}. We use human evaluation to verify the quality of our automatic annotation.
% \heng{In figure 1, I'm not sure what you mean by GPT-4 annotation, and the right part's illustration is confusing, unclear what's going on in "Find summary sentences" and "Find original samples"}
% \daniter{The above paragraphs feels a bit too abstract. The reader doesn't understand the task yet so mentioning gpt annotation is confusing. Perhaps, instead mention an overview of the process (we take sharegpt and use gpt to label each example with a manually created ontology of task types. We use human evaluation to verify the quality of our automatic labeling. Lastly, we report [state some finding], which we obtain by [state some methodology]. }

\subsection{ShareGPT}
ShareGPT is a publically available large-scale collection of user-GPT conversation histories\footnote{The collection of ShareGPT already contains 305,000 samples to the date of this submission, and is dynamically expanding. Since all the data are not publically available, we use the portion that could be downloaded from Huggingface.}. It is based on a Chrome Extension\footnote{\url{https://github.com/domeccleston/sharegpt}} where users can choose to upload their interactions with GPT. The version we used contains $94,145$ split user-GPT conversations and is previously used to train LLMs such as Vicuna~\cite{vicuna2023}. 
Every sample in ShareGPT is a multi-turn conversation with utterances from both users and the GPT engine.
% Every sample in ShareGPT is a multi-turn conversation $C=\{T^i\}$ where $T^i$ denotes the $i$-th turn in the conversation and $T^i$ is composed of user input $U_{user}$ and GPT response $U_{GPT}$. 

\subsection{Self-demonstrated annotation}

The goal of annotating each user query is twofold: to identify the underlying task a user is intending to perform (task types), and to understand the subject matter or field (domains) the query pertains to. 
The annotation process is challenging because i) the annotation requires a comprehensive and suitable predefined set of task types and domains/topics, and ii) the annotation should accurately reflect the genuine requirements expressed in user queries. 
%In this part, we detail our process of self-demonstrated annotation with GPT-4.
We chose to employ GPT-4 to conduct a self-demonstrated annotation due to its superior precision and coverage. %, a critical attribute required for ensuring accurate annotation compared with ChatGPT~\cite{ouyang2022training}. 
The annotation process consists of three stages: 1) chain-of-thought prompting, 2) demonstration sampling, and 3) demonstration pool expansion.
% \heng{write it more clearly on what kind of annotations are you aiming to produce? task types and domains?}

\paragraph{Chain-of-thought prompting.} 

Chain-of-thought (CoT)~\cite{wei2022chain} uses intermediate steps for text generation and improves the performance of LLMs~\cite{chen2023you}. To augment GPT-4's proficiency in comprehending and interpreting user queries, we craft our instructions in manual-CoT style, compelling the LLM to deliver the requisite information in a methodical, step-by-step manner. Specifically, we first ask LLM to identify the domain or topic related to the user query. Then, the LLM is prompted to generate a concise one-sentence summary for the given user query as a reference. Finally, drawing on the insights obtained thus far, the LLM is capable of devising creative and accurate task types corresponding to the user query. 
The generated task types are fine-grained and diverse, spanning from email editing to dream analysis. Overall, we obtain $13,783$ task types and $8,392$ domain labels for all samples in ShareGPT.
% For a complete view of the CoT instruction, please refer to Figure~\ref{framework}(a).
% \daniter{Provide number of task types and domains to highlight how fine grained the labels are.}
% \heng{maybe also give some examples about task types}

\paragraph{Demonstration sampling.}
While the CoT prompting can generate reasonable annotation, it is known that in-context demonstrations can further improve LLM output quality~\cite{wang2022towards}. Thus, we select examples from CoT outputs to serve as demonstrations in the second stage.
%In the second stage, we need to give demonstrations to further enforce the output format of GPT-4 as in-context examples. 
We initiate the demonstration pool with $20$ samples of different domains or topics and task types. For every sample, we randomly select $k$ demonstrations from the pool and append them to the instruction in the first stage. The input and output format could be found in Figure~\ref{framework}.

\paragraph{Demonstration pool expansion.}

To encourage diversity and avoid potential bias in demonstration selection~\cite{selfinstruct}, we gradually expand the demonstration pool. Since we are asking GPT-4 to generate free-form task types, one challenge here is to avoid generating too divergent task types. Therefore, we maintain a dictionary to document the time of appearance for every task type. If a task type appears more than a fixed ratio $\lambda$ among all the current samples, we then add the current sample containing the task type into the demonstration pool. 
% \cz{is this step done during second stage, or after a complete epoch is annotated.} 
By enforcing such constraints, the generated free-form task types could be better ``clustered'' for further analysis.

\paragraph{Experiment settings.} We download ShareGPT from Huggingface\footnote{\url{https://huggingface.co/datasets/anon8231489123/ShareGPT_Vicuna_unfiltered}}, where the 51k conversations are split into 94k ones due to the length limit for input. For every step, $k$ is set to $3$ and $\lambda$ is $0.05$. We concatenate $3$ samples together and let GPT-4 generate the annotation at once for the balance of speed and quality. To encourage diversity, we set the temperature to 0.4 and it takes around 10 days due to speed limitations in GPT-4 to annotate all the 94k samples in ShareGPT. We plan to release all the annotated results for future related research.

\begin{figure*}[!t]
\centering
\includegraphics[width=0.98\textwidth]{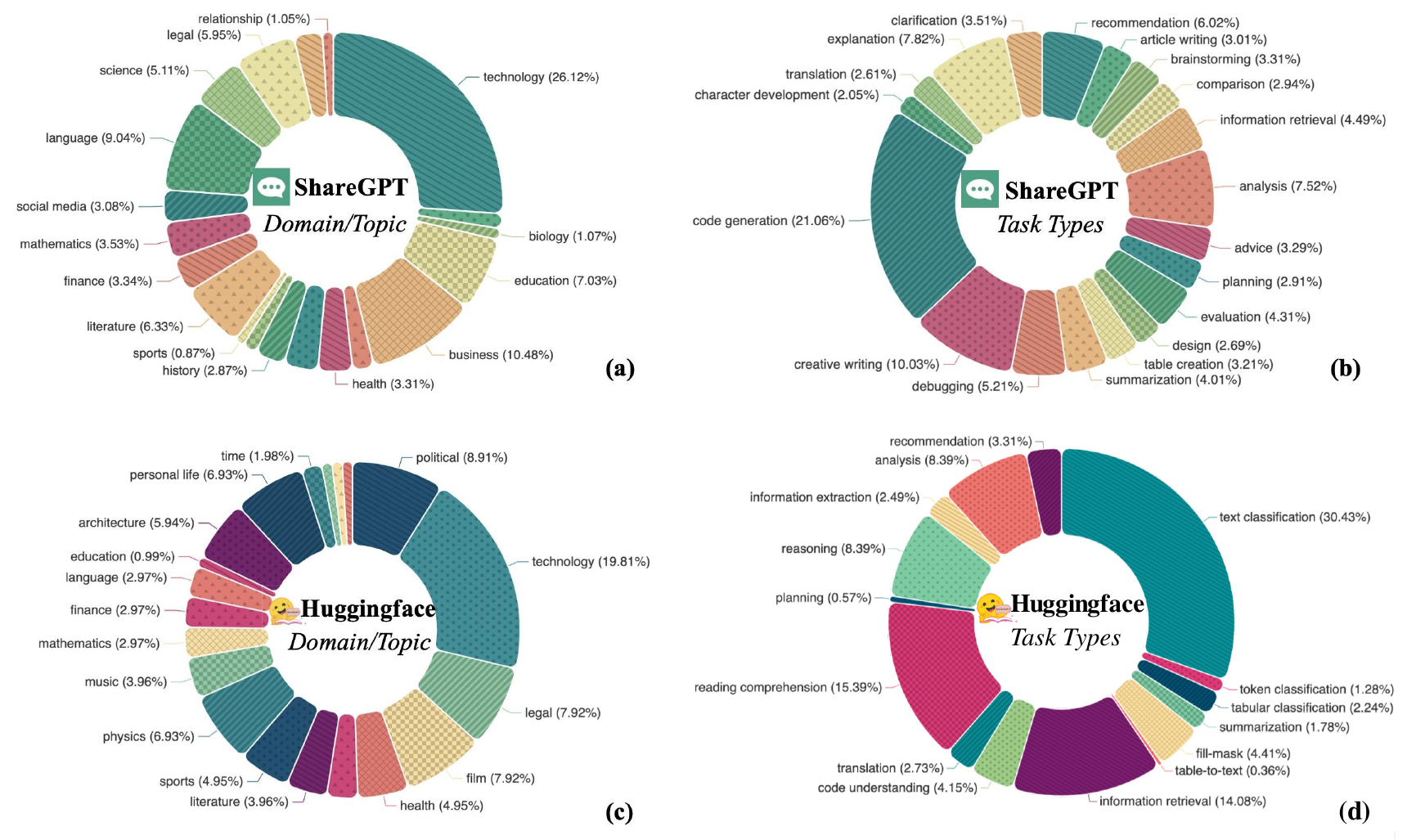}
\caption{Domain  and task types  distributions for ShareGPT user query and Huggingface data.}
\label{fig:domain}
\vspace*{-3mm}
\end{figure*}

\subsection{Human Evaluation}
To assess the quality of annotation produced by GPT-4, a human evaluation is conducted with a specific focus on the generated free-form task types.

We designed and distributed our human assessment task with Doccano\footnote{\url{https://github.com/doccano/doccano}}. We recruited 3 graduate students as our human assessors (all of which are paid as research assistants). The assessors all have rich experiences with related NLP and ML research but were not involved in the development of our framework. We randomly selected 100 samples for evaluation. For every sample, we ask the assessors to judge the generated task types in terms of \textit{completeness} and \textit{correctness}. This is to evaluate whether the generated task types are complete and faithful to the original user query. For \textit{completeness}, the scoring scale is $0$ (bad) and $1$ (good), and for \textit{correctness}, the scoring scale is $0$ (incorrect), $1$ (partly correct) and $2$ (correct). The detailed rubric and the interface are shown in Appendix~\ref{sec:human_eval}.

\begin{table}[t]
\small
\caption{Human evaluation results in terms of completeness and correctness.}
\vspace{-1mm}
\centering
\begin{tabular}{lcc}
\toprule
          & Completeness & Correctness \\
\midrule
Score     &    0.95     &     1.76   \\ \midrule
$\kappa$ &     0.96  &     0.83   \\ \bottomrule
\end{tabular}
\label{tab:human_eval}
\vspace{-1em}
\end{table}

Table~\ref{tab:human_eval} demonstrates the results of human evaluation. We can see that with GPT-4 we got reliable annotations for ShareGPT. Notably, none of the 100 samples got ``incorrect'' annotations. Apart from the scores, we also calculate Fleiss kappa $\kappa$~\cite{Fleiss1971MeasuringNS} for each metric, both indicating ``almost perfect agreement''.

\subsection{Post-processing for analysis}
% \yang{I feel this sub-section can be simplified into one or two sentences. Can we just say we cluster them with KNN based on some heuristic features? And provide more details in Appendix.}

As the domain/topic and task type annotations generated by GPT-4 are free-form words and phrases, clustering the samples poses a significant challenge. For example, ``recipe suggestions'', ``cooking tips'' and ``cooking advice'' all belong to the same task type. To tackle this challenge, we propose a post-processing framework that incorporates three stages shown in Figure~\ref{framework}: (1) a statistical calculation based on heuristic rules, (2) an ensemble with the embedding similarity of summary sentences, and (3) a manual checking process to ensure the best possible quality. Detailed narrations could be found in Appendix~\ref{sec:post_processing}.

\section{Overall Investigation}\label{sec:RQ1}

In this section, we first present the overall statistics and analysis of the ShareGPT clustering results. 
Then we compare the results with conventional NLP datasets.

For conventional NLP datasets, we investigate 2,911 datasets from the Huggingface Datasets\footnote{\url{https://huggingface.co/datasets?p=0&sort=downloads}} collected by~\citet{yin2023dynosaur}. These 2,911 datasets are filtered and selected from the original around 50k datasets in the Huggingface platform with the following conditions: (i) Datasets with no license description or the license does not allow usage. For ethical considerations, the datasets collected are strictly following license restrictions. The number of all datasets fulfilling license requirements is restricted to around 16k. (ii) Datasets that are non-English. Since our investigation focuses on the English language and ShareGPT contains only English samples, datasets collected from Huggingface also follow this paradigm. (iii) Datasets that are related to multi-modal. There is a large portion of datasets in the Huggingface data platform related to multi-modal research. Since we target at NLP community and related topics, we exclude those datasets. (iv) Datasets that are invalid. Some datasets in the Huggingface data platform are empty or cannot be downloaded, which is ignored by our work.

\subsection{Domain and Task Distribution}
Based on our annotation and clustering results, We plot domain and task type distributions of ShareGPT queries in Figure~\ref{fig:domain} (a) and (b). ``Technology'' shares the largest portion of ShareGPT's domain, comprising around a quarter. Other notable domains ``education'', ``business'' and ``language'' make up another quarter. For task types in ShareGPT, around one-fifth is about ``code generation'', which corresponds to ``technology'' in domain analysis. ``Creative writing'' is the second largest portion. The rest of the task types are quite diverse, composing around 2/3 of the whole set. 
% \cz{if these two tasks are underrepresented in Huggingface, let's highlight this point}

% \jh{Overall, the discussion is quite interesting.  However, I feel there are many tasks that actually need sophisticated problem solving instead of NLP or NLG. For example, planning, writing, advising, designing, etc.  I feel if the domain expert can work out an outline, NLG follows such a major line to generate text, the result could be much more interesting and useful.  Quite a portion of novel tasks could belong to this category.  Some deeper insight on distinguishing those that just need some simple correlation on the previous know knowledge from those that need deeper problem solving package as backbone of the answers could lead to better insight in Sections 4 and 5. }

 In the following, we analyze the two mostly seen tasks in ShareGPT, coding and writing assistance, representing $19.9\%$ and $21.3\%$ respectively. 

\paragraph{Coding assistance}

Pre-trained models for programming language have been widely explored in the NLP community~\cite{chen2021evaluating, nijkamp2022codegen}. Current benchmarks used for evaluation~\cite{hendrycks2021measuring} are usually in the form of function generation or code completion given certain requirements. More specific task settings include code search~\cite{husain2019codesearchnet}, code translation~\cite{chen2018tree}, code clone detection~\cite{6976121}, and code refinement~\cite{tufano2019empirical}. We do observe user queries that are similar to the previously mentioned task settings in ShareGPT, such as code generation ($18.6\%$) and code debugging ($9.2\%$). This reflects that the current coding benchmarks are fitting with real-world scenarios. 
However, we still notice a non-negligible portion of requests involving higher-level program understanding, such as code simplification and providing design pattern suggestions, which are seldom captured in existing task definitions.
% \jh{I am also wondering whether LLM can write rather sophisticate and challenging modules. I do not have good knowledge on it.}
% We are also curious about what are the frequencies for some prevalent programming languages. We notice that the majority of the programming language used is Python. Java, JavaScript, CSS, and HTML together amount equally to Python. 
We also plot the proportion of the top 10 most frequent programming languages used in Figure~\ref{fig:pl}. 

\paragraph{Writing assistance}

With advancements in the NLP field, writing assistance tools have shown potential beyond grammatical~\cite{ng-etal-2014-conll} and stylistic improvements, now providing aid in content creation and organization. Writing tasks such as story generation~\cite{fan2018hierarchical} and style transformation~\cite{shen2017style} are popularly explored in the community. Our analysis of ShareGPT usage confirms this trend. For instance, assistance in article drafting and editing accounts up to $5.1\%$ of the writing assistance requests. Similarly, email editing makes up to $2.6\%$ of the queries. These suggest users rely on AI tools for professional writing communication. 

% \jh{Writing could be similar to rewriting some existing passages, I am wondering if GPT can do any creative writing although the term "creative" could be vague and arguable but I largely mean any "creative ideas inside."}  

Despite this, we notice a trend of creative writing for a bunch of text formats, spanning from slogans to tutorial writing. Instead of leveraging LLMs to generate everything, we found a noticeable portion of ``procedure writing'' and ``how-to-write'' queries, underscoring the importance of explanatory and pedagogical writing aids.

\subsection{Distribution Difference with Conventional Datasets}

To provide a comparison of ShareGPT queries with conventional NLP datasets,we also annotate and cluster the collected 2,911 Huggingface datasets in the same way and present results in Figure~\ref{fig:domain} (c) and (d).
We then interpret the differences from two aspects, domain and task type.

\paragraph{Domain Shift} We plot the top 20 most common and representative domains of ShareGPT in Figure~\ref{fig:domain}, which account for around $73\%$ of the entire set. We observe that the domain distribution in ShareGPT is quite diverse. Notably, there is a large portion of technology-related topics. Commonly seen cases involve code writing and debugging. For Huggingface datasets, we randomly select 10 samples from each dataset as an approximation, and leverage GPT to classify the corresponding domain/topics. Similar to ShareGPT, technology-related domains also share a large portion in Huggingface. However, in Huggingface, political, legal, personal life, and film altogether contribute a large portion. We further look into the data source for Huggingface. 
We collect the information in metadata ``Dataset Card'', where detailed information of the specific dataset is introduced. For example, MRPC dataset~\cite{dolan-brockett-2005-automatically} has the statement ``a corpus of sentence pairs automatically extracted from online \textbf{news} sources''. We then use GPT to annotate the corresponding data source as an approximation. We find that the majority of datasets are from Wikipedia and news, occupying a portion of over $80\%$. The rest mainly come from ``government reports'' and ``QA forums''.
This phenomenon calls for dataset curation specific to domains commonly seen in user queries, and the data source should also incorporate materials in multiple formats apart from Wikipedia and news.

\paragraph{Task Type Shift}

% \begin{figure}[hbt]
% \centering
% \includegraphics[width=0.48\textwidth]{figures/huggingface_datasets.pdf}
% \caption{Representative task types of and the corresponding percentage in Huggingface datasets.}
% \label{fig:task_type}
% \vspace*{-3mm}
% \end{figure}

We categorize different tasks based on metadata statistics for all the datasets from Huggingface, as shown on the right side in Figure~\ref{fig:domain}. We filter out tasks with numbers less than $10$, and the rest samples make up to $99.8\%$ of the original set. We observe that question answering and text classification are the top-2 task types, summing up to more than 2/3 of the whole collection. On the contrary, in ShareGPT, almost all the user queries are free-form text generations or generations in a user-required manner.

% \section{Interesting but overlooked tasks discovered in ShareGPT}
% \section{Intriguing  but Overlooked Tasks}
\section{Shifted and Overlooked Tasks}
\label{sec:RQ2}

\begin{table*}[!ht]
\caption{Summary of the long-tail tasks discovered. ``\textcolor{red}{||}'' indicates concatenated user queries in the same conversation.}
\vspace{-0.5em}
\centering
% \small
\resizebox{\textwidth}{!}{
\begin{tabular}{llll}
\toprule
Task type & Characteristic & Proportion & Example \\ 
\midrule
\multirow{4}*{Advice}&Offering tailored&\multirow{4}*{$3\%$} & My dad, he's a rude person. he doesn't like me, he makes it clear. he calls  \\
~&acvice under&&me bad names, he has threatened to kick me out. should I be concerned? \\
~&given scenarios&&considering people have suffered through worse \textcolor{red}{||}  I need money but I can't\\
~&&& get it because working at my dad's pub is too much for my autism \\
\midrule
\multirow{4}*{Design}&Construction of&\multirow{4}*{$2.5\%$} & Can you help me think of some logo ideas for ``StartUpScout``? \textcolor{red}{||} Can you give  \\
~&some objects or && more similar ideas like ``A playful cartoon scout character holding a tech startup \\
~&for implementation&& icon''? \textcolor{red}{||} I liked this idea: A cartoon scout character with binoculars and a backpack, \\
~&of an activity && walking towards a futuristic city skyline in the distance. Can you tell me more details? \\
\midrule
\multirow{5}*{Planning}&Providing a &\multirow{5}*{$2.7\%$} & I'm going on a road trip with my daughter on 3rd April. We will be leaving\\
~&sequence of&&London and want to be in Nice, France on 9th or 10th April where we'll stay \\
~&steps to achieve&& about 3 days. We then need to be in Calais on 15th or 16th to take the Eurotunnel. \\
~&a pre-defined&&Everything else is undecided. Can you give a plan for routes where we'll be doing \\
~&objective.&&no more than 5 hours driving per day. We have an EV and need to stop for charging. \\
\midrule
\multirow{3}*{Discussion}&Exchanging views&\multirow{3}*{$3.8\%$} & Should we help local charities or overseas charities? \textcolor{red}{||} Some people \\
~&or ideas on a && think that being rich and famous leads to happiness. There is also a saying  \\
~&specific topic&&``Money can buy comfort, but it cannot buy happiness.'' What do you think?\\
\midrule
\multirow{3}*{Analysis}&Examination of&\multirow{3}*{$7.3\%$} & I need some help analyzing a poem that I've been reading for my English Literature   \\
~&a target for its&& class. The poem is "The Road Not Taken" by Robert Frost. \textcolor{red}{||} Thank you, that's a great \\
~&nature and structure&&start. Could you help me analyze how Frost uses metaphor in this poem? \\
\midrule
\multirow{5}*{Evaluation} &Determination&\multirow{5}*{$4\%$}&I want you act as a resume evaluator. Here are the rules: [Format]: Use a clear, \\
~&of the subject's,&&easy-to-read font and suitable layout ... [Content]: Your resume should include...\\
~&properties, based&&  You should be able to read example url and anlyze it. Here are some good example \\
~&on given rubrics&&for resume: <https://docs.google.com/...> Each section means 20 point, \\
~&&&total 100 points. Just show me explaination and points then organize to table. \\
\bottomrule
\end{tabular}
}
\label{tab:long_tail_example}
\vspace*{-4mm}
\end{table*}

In this section, we detail the overlooked tasks discovered in the analysis process of ShareGPT, with concrete examples shown in Table~\ref{tab:long_tail_example}. Task selection is based on the distribution in Figure~
\ref{fig:domain} that satisfies two criteria: (1) long-tail tasks of the distribution, summing towards around 40\%; and (2) tasks that are not overly skewed in the distribution, constituting around 25\% - a comparable figure to the predominant segments. We also summarize the features for each task type in Table~\ref{tab:long_tail_summary}, along with the potential roadmap. Commonly seen topics and human analysis for performance of LLMs are shown in Figure~\ref{fig:long-tail} in Appendix~\ref{sec:topic_task}.

\subsection{Task of Providing Advice}

The task of giving advice occupies a portion of up to $3\%$ in ShareGPT. 
The concept of a machine offering advice has been present and associated with NLP from as early as the 1970s~\cite{shortliffe1973artificial} with the development of expert
systems~\cite{liao2005expert}. At that time, giving advice is usually performed with a consultation program~\cite{scott1977explanation}, which plays the role of an expert consultant in restricted domains, such as health. Since then, the field has evolved significantly, with expanding domains into legal~\cite{pinkwart2006toward}, finance~\cite{radford2003practice}, etc. The advising process back then is more like a QA system, where the user first input background confined by rules, together with a direct question such as ``Is Organism-1 a rod or coccus (etc.)?''. Later, advice was formally characterized as a multi-step process involving the analysis of the recipient's context and the formulation of a response in natural languages~\cite {saeidi2018interpretation}. 

Compared with previous advising systems that require rigid rules and were not designed for open-ended dialogue, user queries in ShareGPT are  more free-form.
% As seen in Table~\ref{tab:long_tail_example} for an emotional support situation, when users solicit advice, their prompts are often mixed up with current scenarios, user intents, and advice requests. 
Moreover, instead of focusing on restricted professional  domains, these requests are more akin to everyday tasks, such as relationship dilemmas and personal careers. 
This also presents the requests of more general, macro-level guidance, compared with  providing micro-specific answers in previous studies. 
% As such, the tasks require proficiency in user profile/intent acquisition, personalization, and broad domain knowledge.

% There are many challenges in this task category. \cz{another challenge comes from evaluation} First and foremost, the model needs to have an accurate view of user scenarios. When expressing their queries, we often observe informal, unstructured, and even sentimental statements about a very complicated situation. \heng{When NLP tasks were proposed in the past, the people who proposed the tasks were all NLP experts, so they know what tasks are suitable / feasible for the current NLP technologies. For example, they know how models like GPT work, and they are designed for NLG task, so they will not expect ChatGPT to work like a therapist or problem solver. So I think when we advocate that the community to work on certain types of new tasks, we need to point out the additional components/techniques that need to be used on top of LLMs} Also, it could be difficult for models to provide personalized advice, especially for topics that are intrinsically subject to individual variation. 
% Therefore, future research could concentrate on advancing LLMs' understanding and capabilities in delivering customized advice, respecting the user's individuality and the context at hand. Furthermore, ethical considerations are crucial in this task category, underscoring the need for future work to also focus on promoting responsible and ethically sound advice-giving models.

\subsection{Task of Designing}
Request to design some objects with specifications constitute $2.5\%$ of ShareGPT queries. The task of designing progresses slowly in the field of NLP over the past half-century, and does not have a clear line of development. Previous NLP-related works generally focused on relatively narrow domain-specific topics, such as entity-relationship schema~\cite{habib2019automated} and user interface~\cite{sontakke2014rule}. 

While previous works focus more on the design of structured objects, such as database UI, user queries arising from ShareGPT usually seek designs for more complicated objects, spanning from questionnaires to websites. This indicates a trend that the scope of design has broadened to include not only the structure of tangible objects but also conceptual frameworks and processes. 
Another significant difference is that design-related queries in ShareGPT ask for more diverse and creative requirements, especially for personalized objects such as icons.
Additionally, in ShareGPT, we see instances where users demand designs with very specific constraints, such as adherence to a particular color scheme in interface design. 
These ``visual image design'' tasks, though communicating in texts, require the model to have a sense of visual understanding in order to generate aesthetically pleasing outcomes. 

\begin{table*}[!ht]
\caption{Features of the tasks discovered and the potential roadmap. 
}
\vspace{-0.5em}
\centering
% \small
\resizebox{\textwidth}{!}{
\begin{tabular}{llll}
\toprule
\textbf{Task type} & \textbf{Before LLM}  & \textbf{After LLM} & \textbf{Roadmap} \\ 
\midrule
\multirow{2}*{Advice}&rule-based; limited context analysis;  & free-form; open-ended dialogue;&emotion perceivable;\\
~&professional domains; micro-specific answers& general everyday tasks; macro-level guidance&personalization;\\
\midrule
\multirow{3}*{Design}& standardized requirement; &creative design; user-based constraints;&multi-modality;\\
~&tangible and structured objects;& expansion to conceptual frameworks and processes; &interactivity;\\
~&one-off, static design;& interactive design with feedback;\\
\midrule
\multirow{2}*{Planning}&formal/open-form language; context-insensitive; & free-form language; user-context understanding;&better reasoning;\\
~&micro-level actions in specific domains; & macro-level planning for many aspects in daily life;&world-knowledge\\
\midrule
\multirow{2}*{Discussion}& non-interactive with pre-defined inputs; & highly interactive, reacting dynamically;&personalization;\\
~&structured around specific domains;& encompass broader subjects; personalized with empathy&interactivity;\\
\midrule
\multirow{2}*{Analysis}& mostly classification; pre-defined aspects; & free-form input; mostly unspecified targets;&multi-modality;\\
~&limited domains focused;& wider analysis scope;&better reasoning;\\
\midrule
\multirow{2}*{Evaluation}& mostly plain texts for evaluation;& much diverse input formats; &fairness;\\
~&metric designed for specific tasks;&human-aligned, context-specific, open-ended metrics; &personalization;\\
\bottomrule
\end{tabular}
}
\label{tab:long_tail_summary}
\vspace*{-4mm}
\end{table*}

\subsection{Task of Planning}
% \heng{LLMs are very good at connecting dots / finding associations among words/concepts, so there is some certain creativity / incremental improvement based on existing design/planning. If we are advocating this to be a new NLP task to focus on, we need to explain why LLMs can be naturally used as a starting point}
Planning is another important task we identified, which constitutes approximately $2.7\%$ in ShareGPT. 
% It is a systematic and strategic process that involves establishing a sequence of steps to achieve a pre-defined objective. 
Planning in NLP has a long research history. Back in $1969$, PLANNER~\citep{10.5555/1624562.1624592} was invented as a language for manipulating models in a robot. Follow-up works \cite{10.5555/889207, bonczek1979computer} mostly focus on planning with programming languages for decision-making. Later, a plan was formally defined as an assembly of operators~\cite{grosz1988plans} that describes the transition from initial states to final states, where rules and schema were designed for induction. Afterward, planning was employed in different domains and scenarios, such as trajectory~\cite{borrelli2006milp}, autonomous agents~\cite{chen2009integrating}, and power dispatch~\cite{estevam2010reactive}. Most of the works focus on planning with formal languages under certain rules. 
Nowadays, many benchmarks related to planning have emerged~\cite{valmeekam2022large, xie2023translating,schema2023e}. Although they require planning with open-form natural languages, they mostly focus on rudimentary tasks with simple actions such as (e.g., ``put up a block'' for ``arrange blocks'')~\cite{valmeekam2023planning}.

The emergence of LLMs has spawned much more free-form and customized formats in planning. One  example is illustrated in Table~\ref{tab:long_tail_example}, where users ask for planning a route with specific constraints on time, places, and EV charging. We noticed a trend of macro planning, e.g., travel plans, and course plans, instead of planning for micro actions in previous NLP studies. The domains entailed also greatly widen, spreading to many aspects of everyday lives, compared to previous planning systems designed for specific functions/users. Therefore, these planning tasks usually require a higher level ability in personalization, reasoning, and knowledge integration, where follow-up research efforts could lay hands on.
% To illustrate, in the process of generating travel plans, the model must possess a commonsense understanding of related information, such as destination and the user's preferences and time constraints. It should harness both mathematical and temporal reasoning to rule out specific scheduling, thereby enhancing the feasibility of the plan. In ShareGPT, we also observe cases when users pose specific constraints and requirements, e.g., avoidance/preferences of certain places in a travel plan. 

\subsection{Task of Discussion}

Discussion is an interactive and dynamic exchange of ideas or viewpoints, which consists of $3.8\%$ samples in ShareGPT. The act of discussion in NLP is mostly explored in the form of conversation generation with chatbots~\cite{goldenberg1992instructional}, and they mainly focus on structured discussions in specific domains that require domain expertise, such as political debating~\cite{mirkin2017recorded}. Another notable characteristic is the non-interactivity in many previous works~\cite {DBLP:conf/acl/ZhangLPGC19, ouyang2020dialogue}, although they promote multi-turn generation~\cite{DBLP:conf/lrec/ChenLSYZWHZ20} for several fixed user inputs.

However, user queries in ShareGPT are typically more dynamic and unpredictable, encompassing a vast array of subjects and requiring a deep understanding of various perspectives. For instance, in ShareGPT, there are cases when users initiate philosophical discussions such as ``What's the meaning of life?'', which may require self-thinking in viewpoints. On the other hand, the discussion process in ShareGPT is quite interactive, which poses challenges in actively taking initiative and even shifting topics. Developing models with certain personalities and empathy to facilitate more effective and meaningful discussions will be helpful.

\subsection{Task of Analysis}

The task of analysis takes up a very large portion of ShareGPT, approximately $7.3\%$. Textual analysis is a long-standing and crucial branch of NLP. In the early stages, researchers focus on analyzing linguistic features within texts, such as syntax~\cite{floyd1963syntactic} and discourse~\cite{harris1970discourse}. Gradually, they began to wonder ``how can analysis of the patterns of words and grammar contribute to an understanding of the meaning''~\cite{stubbs1996text}, accompanied with the investigation in both directions of semantic/contextualized analysis~\cite{mann1988rhetorical, fairclough1992discourse} and larger-scale texts. As textual analysis evolves and improves, they are gradually employed in various domains like social science~\cite{fairclough2003analysing}, medical~\cite{edara2023sentiment}, and finance~\cite{fisher2016natural} with different objectives like culture~\cite{carley1994extracting}, sentiment~\cite{nasukawa2003sentiment} and opinions~\cite{cambria2013new}. 

% Traditional analysis tasks~\cite{medhat2014sentiment, brown1983discourse} in NLP differ greatly from real-world scenarios. The instances for analysis are usually commonly seen in real life, such as film, contracts, and even political events. 
% \heng{again the difference is not about 'real-world' or not. The target users for traditional NLP tasks are often for government users/analysts for example, those scenarios are also real-world. I think the difference is between a specific group of users from funders vs. any regular users who need help on their daily tasks}

Even though previous research has already covered many fields and objectives, we still observe striking differences when it comes to user queries in ShareGPT. Notably, many previous analysis tasks take the form of classification, e.g., identifying a risk type in a financial report with $25$ total risk types as label space~\cite{loughran2020textual}. User queries, in contrast, usually do not specify the targets/aspects for analysis, as exemplified by the literature analysis case in Table~\ref{tab:long_tail_example}. The scope of analysis also varies greatly in ShareGPT, ranging from the overall analysis of classical music development to the functional analysis of a single function in code. Hence, it calls for better specifications for user requirements/intents, as well as customization to different levels of scope.

\subsection{Task of Evaluation}

In ShareGPT, ``evaluation'' queries constitute up to $4\%$. The concept of evaluation permeates almost every aspect of NLP. Standard metrics such as F1 measure~\cite{chinchor-1992-muc}, ROUGE~\cite{lin-2004-rouge}, and BERTScore~\cite{zhang2019bertscore} are mostly employed to evaluate classification or generation results at the sentence level.
Recently, there is a  surge in research efforts to improve alignment with human preferences~\cite{zhong2022towards, liu2023gpteval, fu2023gptscore, luo2023chatgpt} by using larger models. 
% Despite the sentence-level evaluation metrics, document-level metrics~\cite{jiang2021blonde} are also designed, mostly in the domain of machine translation~\cite{comelles2010document}.

% However, the current evaluation paradigm in NLP is proved to be positional biased~\cite{wang2023large} (i.e., prefer context generated by LLMs), and has a specific preference for contexts with long length~\cite{wang2023far}.

However, evaluation-related queries  from ShareGPT  are quite different. First, we observe that evaluation objects shift from traditional plain texts to a variety of input formats. For instance, GPT is often utilized to assess resumes or code snippets. Furthermore, the evaluation metrics used in these scenarios are exceptionally diverse and open-ended, ranging from the influence of a brand to the feasibility of a plan. This is quite different from previous metrics that are specifically designed for summarization or machine translation.
% Existing NLP analytics can address some aspects, e.g., grammar checking~\cite{soni2018systematic}, but not all aspects in real-world user queries. 
% One potential solution is to strengthen the understanding of different input formats, specifically those structured ones such as code snippets and resumes. Also, instead of requiring LLMs to directly output the evaluation results, it would be helpful to trace back how they come to the results.

% \subsection{Other Tasks}

% Other notable tasks include ``child education (0.8\%)'', where GPT is employed to help summarize, simplify, or explain complicated or advanced concepts to children. Although the application of NLP in education is a long history~\cite{alic2022computationally}, little attention is paid to education for young kids. \heng{I don't think this statement is true. There are many efforts on NLP for young children education}

% GPT is also popularly used for ``Mock test (0.9\%)'', where users ask GPT to generate questions they will probably encounter for their job interviews or exams. 

% \heng{I feel to give a list of tasks is not enough. You should suggest some concrete solutions and possible roadmaps for each task}
\section{Emerging Trends and Challenges}
\label{sec:trends}
In this section, we summarize the common trends and challenges shared by these prevalent or overlooked tasks we identified from the ShareGPT data.

% We summarize our findings and provide insights regarding the trends and some existing challenges of emerging tasks in ShareGPT in this section.

\subsection{What trends are reflected in ShareGPT user queries?}

In view of user queries in ShareGPT, we notice incredible expansions of task scopes and user bases. 
\paragraph{More Aligned with Daily Life}
GPT is leveraged for all kinds of everyday tasks, straddling both professional and personal issues. As such, user queries exhibit an increasing tendency to be free-form and contain arbitrary input, reflecting everyday life scenarios. It is also more customized as a personalized assistant, covering a broad range of topics with nuanced requirements.

\paragraph{Diverse User Groups}
Accompanying the prevalence in everyday tasks is the diversification of user groups. We observe queries by diversifying users of different ages, professions, cultural backgrounds, and even traditionally marginalized groups.

% \paragraph{Why do these trends emerge?}
% growing public awareness and acceptance of LLMs.

% LLMs are within reach

% LLMs are powerful

\subsection{What challenges are proposed by trending and future tasks}
Despite the unprecedented success of LLMs,
we notice real-world users are also raising their requirements when querying an LLM.
% we still observe several aspects that warrant exploration. 
Some of the concrete examples are shown in Appendix~\ref{sec:examples}.
% \yang{Don't say too much about what LLM cannot do unless you have proof or cases for that, talk more around these tasks. What requirements they are asking LLM to have?}
% \vspace{-2mm}
\paragraph{Better Reasoning Capacity}
One evident need that emerged from user queries is advanced reasoning abilities. Users expect LLMs to comprehend complex scenarios, infer causality, and develop well-organized feasible responses to help them, especially with strategic decision-making. 
% \vspace{-2mm}
\paragraph{Emotion Perceivable}
A non-negligible portion of user queries come from marginalized groups seeking help, often due to real-world communication challenges or inadequate solutions. LLMs interacting with these users must effectively perceive their emotions and exhibit empathy, particularly as they may be sensitive when discussing their circumstances, such as those with mental problems. This necessitates a human-centric approach from LLMs, cultivating a safe environment for free expressions of concerns and offering appropriate support.
% \vspace{-2mm}
\paragraph{World Knowledge}
In managing the diversity of user queries pertaining to everyday life, the imperative for LLMs to effectively utilize world knowledge grows apparent. This spans from static factual data, such as intercity distances for road trip planning, to dynamic, evolving information like restaurant recommendations fluctuating with Google ratings. Although integrating external plugins~\cite{schick2023toolformer} and applications is a viable initial strategy, meticulous attention must be paid to uphold the neutrality and accuracy of this knowledge, mitigating potential biases and misinformation.
% As users input diverse queries related to everyday lives in the real world, it becomes important for LLMs to harness world knowledge. For example, the distance between two cities in road trip planning. The expansive knowledge base of LLMs should not only be a repository of factual information but should also capture the dynamics of the evolving world, as requested in restaurant recommendations with Google ratings changing over time. Integration with external plug-ins and applications could be a starting point, but care should be taken to ensure the neutrality and accuracy of the knowledge to avoid potential biases and misinformation.
% \vspace{-2mm}
\paragraph{Multi-modality}

Though restricted to textual interaction with LLMs, user queries demonstrate the potential of a diverse range of modalities. We observe the textual descriptions of images, websites, and UIs, as well as URLs to music/videos in user inputs, which calls for the development of multi-modal integrations.
% \vspace{-1mm}
\paragraph{Personalization and Fairness}
We observe users increasingly expect AI models to understand their unique needs and offer tailored solutions. We also notice the diverse user bases for LLMs of different groups. The drive for personalization must be balanced against the principle of fairness. Personalized responses should not amplify biases or perpetuate unfair outcomes, and the pursuit of fairness should not impede the assistance for customization.

% \vspace{-2mm}
\paragraph{Dialogue and Interaction}
For user queries that entail massive collaborations or discussions with LLMs, they require  a high degree of interactivity, which doesn't just involve passively accepting user queries and providing answers, but actively engaging with users to exchange viewpoints and ideas. This kind of interactive dialogue can help create a more user-friendly experience, facilitating a deeper understanding of the user's needs. 

\section{Conclusion and Future Works}
In this paper, we identified a discrepancy between the existing state of NLP research and the need for real-world applications by investigating large collections of ShareGPT and Huggingface data samples with GPT-4. We make this annotation resource public, which could be directly leveraged for further investigation of ShareGPT data, or to fine-tune advanced models such as Llama as a much cheaper alternative tool for annotation. Based on our observation, we also provide insights into the challenges posed by real user needs and a potential roadmap for future work.

\section*{Limitations}

We discuss the limitations of this work in the following aspects:
\begin{enumerate}
    \vspace{-2mm}
    \item Our study is based on two sources, ShareGPT and Huggingface datasets. Although they are the most abundant resources we can obtain at hand to represent user queries and the traditional benchmarks in the NLP community, they could hardly reflect the whole breadth of real-world situations. Actually, both sets are still growing dynamically as time flows.
    \vspace{-2mm}
    \item In our annotation process, we employed GPT-4, the state-of-the-art LLM to help generate domain/topics and task types. On top of that, we conduct a human evaluation for quality assessment. However, there are still chances when the annotation from GPT-4 is not accurate enough, which could influence the post-processing step.
    \vspace{-2mm}
    \item Continuing from 2, our work relies on the usage of LLMs. We require annotation from GPT-4 for every sample in ShareGPT and the selected set in Huggingface datasets, which is a large number. Despite that, we make the annotation results publicly available, this annotation process is extremely resource-intensive and time-consuming to reproduce. 
\end{enumerate}

\section*{Ethics Statement}

Our work highlights the shifted trend and the overlooked problems of previous NLP studies. By analyzing real-world user queries, we examine the new requirements that emerged and hope to make LLMs more beneficial and better aligned with their users' needs, including the marginalized groups. We hope that our work can be an initial effort to mitigate the gap between user needs and academic benchmarks in the era of LLMs. Overall, we do not foresee any major risks or negative societal impacts of our work. The ShareGPT and Huggingface datasets we experiment with are publicly available online. We have open-sourced this project to facilitate future research, especially for small research groups or institutions with relatively fewer resources of LLMs.

\section*{Acknowledgement}
Research was supported in part by US DARPA KAIROS Program No. FA8750-19-2-1004, National Science Foundation IIS-19-56151, and the Molecule Maker Lab Institute: An AI Research Institutes program supported by NSF under Award No. 2019897. Any opinions, findings, conclusions, or recommendations expressed herein are those of the authors and do not necessarily represent the views, either expressed or implied, of DARPA, the National Science Foundation, or the U.S. Government.

\bibliography{anthology,custom}
\bibliographystyle{acl_natbib}

\newpage
\appendix

\section{Details in Post-processing}
\label{sec:post_processing}

In this section, we detail the framework for post-processing. After annotation was completed by GPT-4, we got 1) free-form words and phrases for domain/topic, 2) a one-sentence summary for the user query, and 3) free-form words and phrases task types.

We start by calculating the frequency of certain words/phrases. Considering the existence of synonyms, such as the ``advice, tip, suggestion'' mentioned before, we resort to an external dictionary\footnote{\url{https://www.thesaurus.com/}} and combine synonyms together. On the other hand, those semantically similar words/phrases should also be clustered together. Previously we require GPT-4 to generate summaries for user queries of great quality, and they could be used as references in finding representative samples for the same cluster. For every sample combined by the previous heuristic rules, we search for their k-nearest-neighbors~\cite{keller1985fuzzy} and union all the samples as the final results. For words/phrases of very low frequency not incorporated above, we select the nearest ``cluster'' as the approximation. Finally, to ensure the best possible quality, we manually filtered out unrelated ones in the processed results. 
% \heng{what is the purpose of clustering words/phrases?}
% \daniter{Why not select a set of words that represent your ontology of labels and then constrain GPT4 output to label each example with the predefined clusters rather than clustering the unconstrained outputs of gpt?}

\section{Human Evaluation Interface}
\label{sec:human_eval}

\begin{figure*}[!t]
\centering
\includegraphics[width=0.98\textwidth]{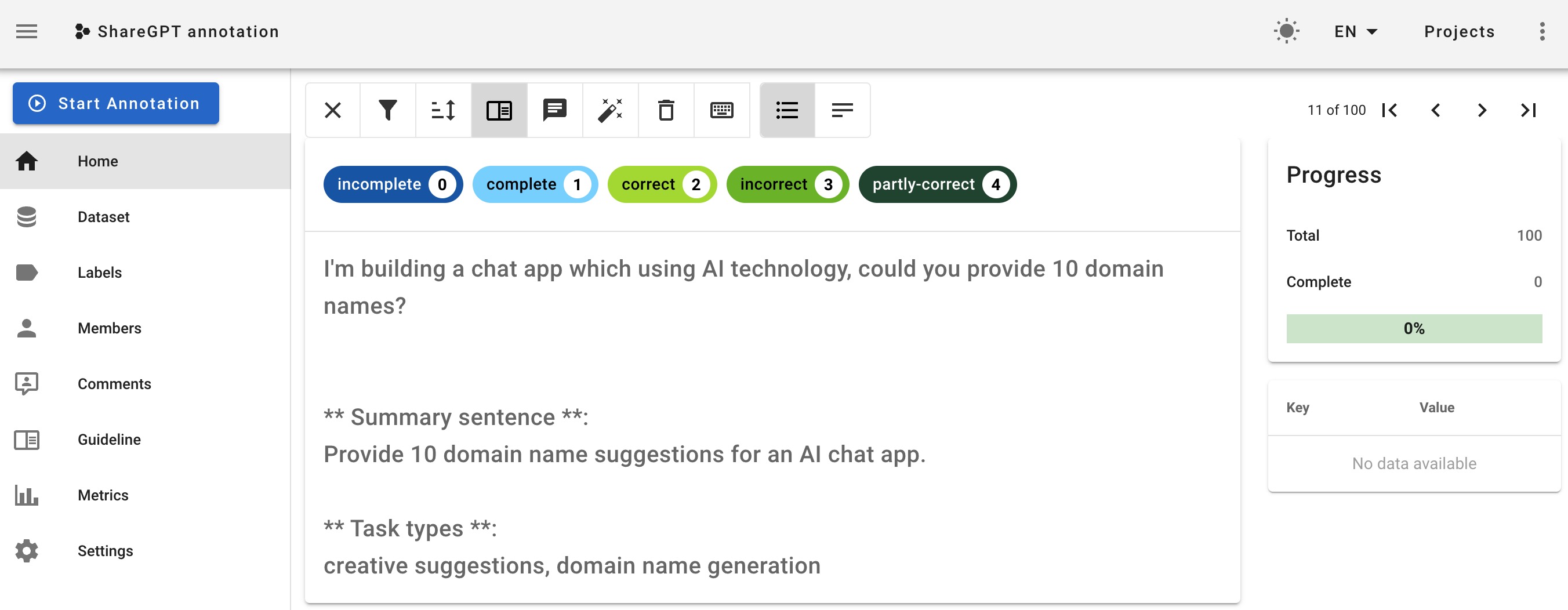}
\caption{The interface for human assessment. The assessor is shown a user query sampled from ShareGPT, the summary sentence of the user query as a reference, and the generated task types labeled by GPT-4.}
\label{fig:human_eval_interface}
\end{figure*}

\begin{figure*}[!t]
\centering
\includegraphics[width=0.98\textwidth]{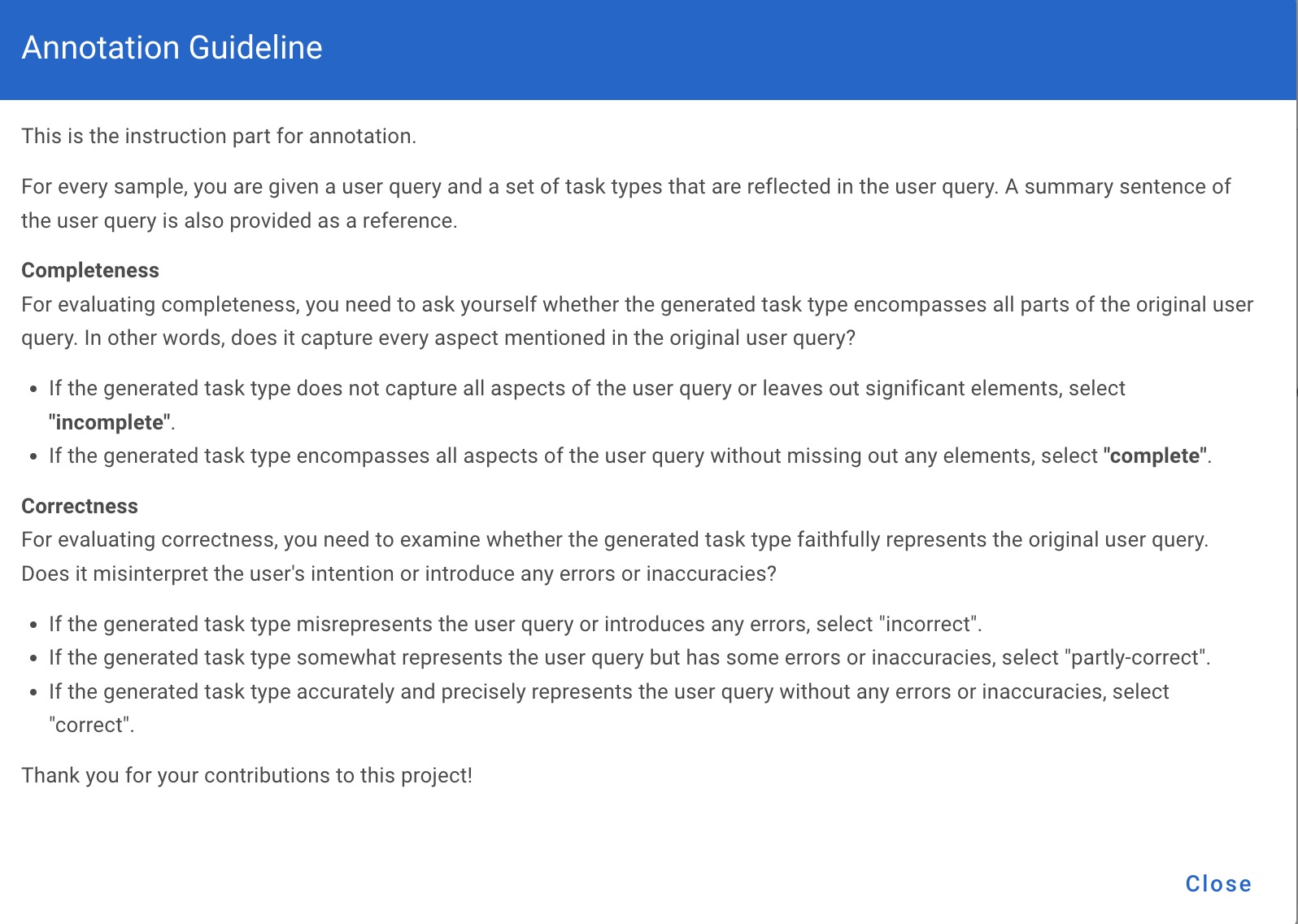}
\caption{The annotation guideline presented to assessors of our human evaluation process.}
\label{fig:human_eval_guideline}
\end{figure*}

The website interface screenshot adapted from Doccano for human evaluation is shown in Figure~\ref{fig:human_eval_interface}. Assessors were informed of the purpose of the study.
Before they began to work on the assessment task, they were presented with task instructions shown in Figure~\ref{fig:human_eval_guideline} and a rating example.

\section{Popular programming languages seen in Section~\ref{sec:RQ1}}
\begin{figure}[h]
\centering
\includegraphics[width=0.48\textwidth]{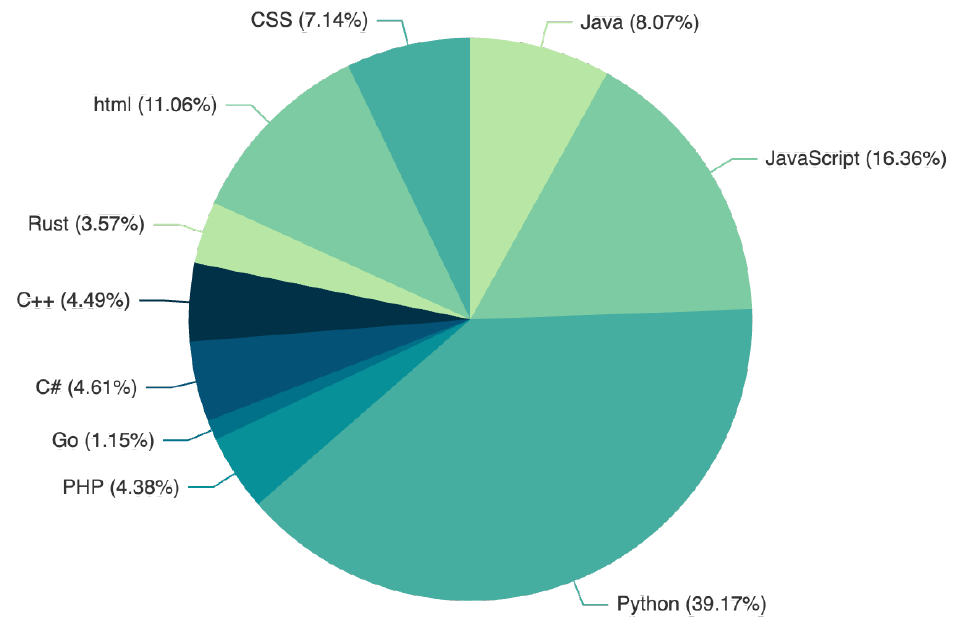}
\caption{The top 10 most commonly used programming languages in ShareGPT.}
\label{fig:pl}
\end{figure}

We summarize the top 10 mostly used programming languages in coding assistance tasks of ShareGPT. As shown in Figure~\ref{fig:pl}, Python is the mostly used programming language. Apart from that, we observe another large portion of HTML and CSS, reflecting great user needs in website design and programming.

\begin{figure*}[!t]
\centering
\includegraphics[width=1\textwidth]{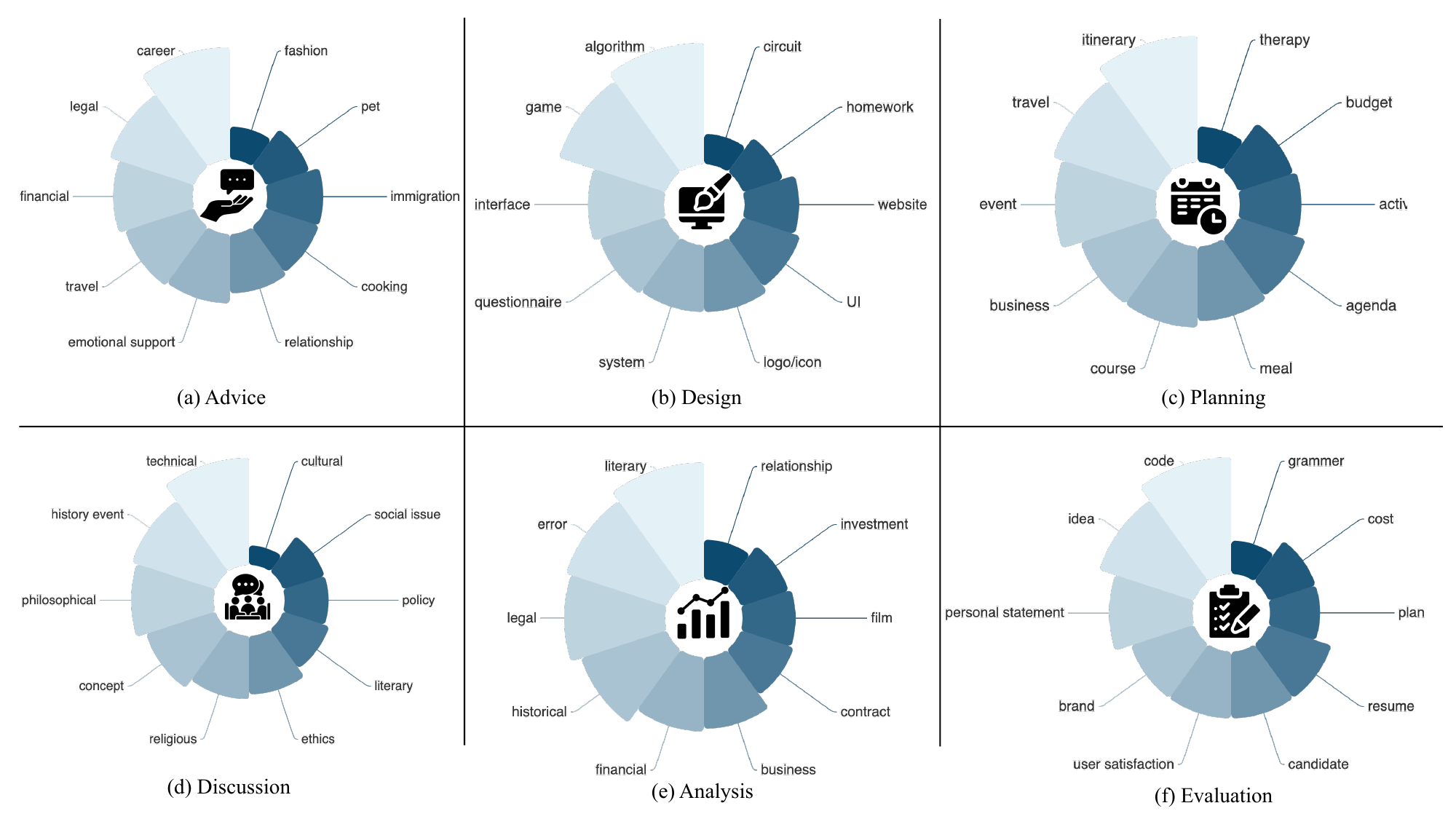}
\caption{The top 10 most commonly seen topics for the novel tasks discovered from ShareGPT. 
% \cz{The angle range and length of each sector is the same for all different tasks. Need to correspond to actual ratio.}
}
\label{fig:long-tail}
\end{figure*}

\begin{table*}[t]
\caption{Case-based analysis for the performance of LLMs.}
\centering
\begin{tabular}{lcccccc}
\toprule
          & advice & planning & design&discussion&analysis&evaluation \\
\midrule
GPT-3.5-turbo     &    0.55     &  0.80&0.70&0.65&0.70&0.75   \\ \midrule
GPT-4 &     0.40&0.60&0.65&0.45&0.50&0.45 \\ \bottomrule
\end{tabular}
\label{tab:case_analysis}
\end{table*}

\section{Common topics and LLM performance for tasks listed in Section~\ref{sec:RQ2}}
\label{sec:topic_task}
We plot the top 10 mostly seen topics discovered for each task type mentioned in Section~\ref{sec:RQ2} shown in Figure~\ref{fig:long-tail}. The shallower color and longer bar indicate a larger portion. For ``advice'', macro-level guidance such as career advice was the most sought-after category, reflecting individuals' focus on professional growth and job success. This was followed by legal and financial advice, highlighting the everyday complexities people face in navigating legal systems and managing their personal finances. For ``design'', apart from algorithm design, we also notice creative design requirements such as interface, game, and website, which may also entail visual perception. In ``planning'', the most common topic discerned was itinerary and travel, indicating significant needs in devising travel plans or daily schedules.
As to ``discussion'', there are explicit topics for discussion such as the historical event or for coding design. We also notice abstract topics like philosophical discussions and ethics discussions, which raises the higher requirement for the self-thinking of LLMs. When it comes to ``analysis'', a broad range of topics are covered, from specific domains such as literary and historical to personal affairs like relationship analysis. Finally for the task of ``evaluation'', we notice a diversity of objects to be evaluated, spanning as concrete as a code snippet or grammar and as abstract as candidates for certain positions.

To help better understand the difficulty of the newly identified tasks, we provide a case-based analysis of LLMs. Specifically, we did a preliminary study by randomly selecting 20 samples for each task type from the ShareGPT data. We manually examined the performance of two models on each case, and reported the failure rate for GPT-4 and GPT-3.5-turbo in Table~\ref{tab:case_analysis} with respect to each task type.

\section{Concrete examples mentioned in Section~\ref{sec:trends}}
\label{sec:examples}

In this part, we display concrete examples in ShareGPT where requirements raised by users pose specific challenges. We highlight the challenging requirements and the misinformation generated by GPT-4 in \textcolor{red}{red}. 

\begin{figure*}[!t]
\centering
\includegraphics[width=0.98\textwidth]{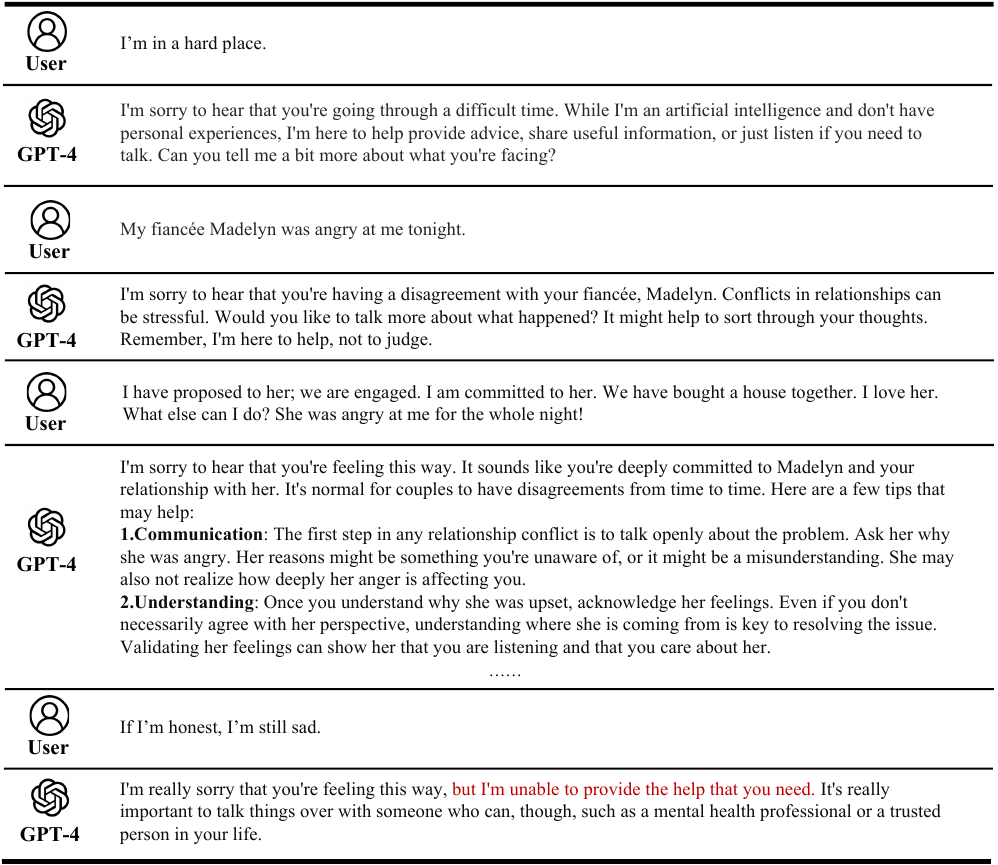}
\caption{An example of relationship counseling and the corresponding GPT response.}
\label{fig:relation_counsel}
\vspace*{-3mm}
\end{figure*}

One example of ``advice'' about relationship counseling is shown in Figure~\ref{fig:relation_counsel}, where the user is seeking emotional support caused by his relationship with his fiancee. We can see that GPT-4 lacks emotional perception, repeating ``I'm sorry to hear that...'' during the whole interaction, and failing to demonstrate empathy towards the user scenario.

\begin{figure*}[!t]
\centering
\includegraphics[width=0.98\textwidth]{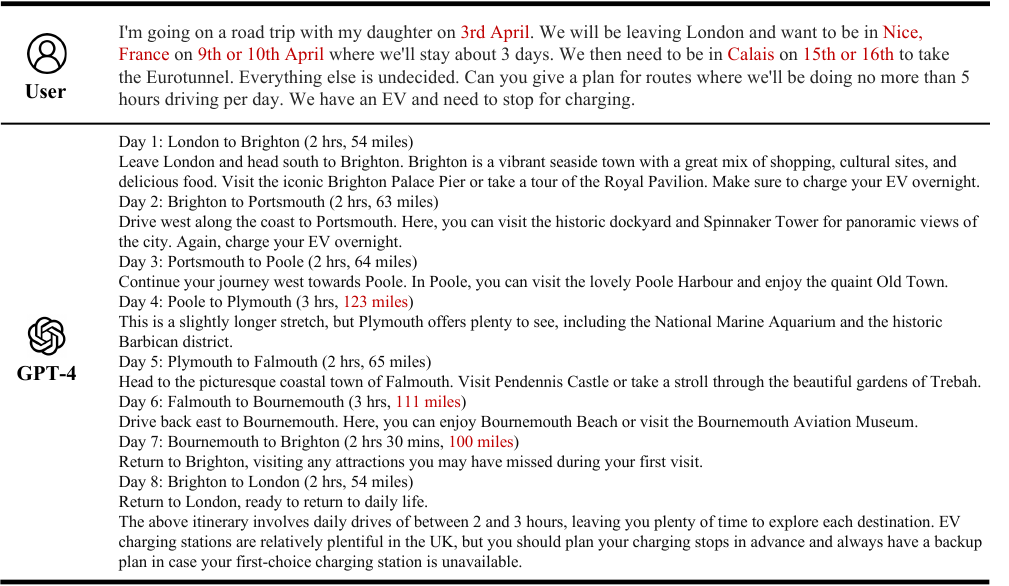}
\caption{An example of route plan and the corresponding GPT response.}
\label{fig:route_plan}
\vspace*{-3mm}
\end{figure*}

In Figure~\ref{fig:route_plan} we display an example of ``planning'', where the user cast specific constraints on time and places, and specific requirements for EV charging. GPT-4 generates an 8-day plan, which does not fulfill the requirement for dates. The route is also fully in the UK, failing to cover places like Nice and Calais. Finally, GPT-4 incorporates misinformation for distances between two cities. 

\begin{figure*}[!t]
\centering
\includegraphics[width=0.98\textwidth]{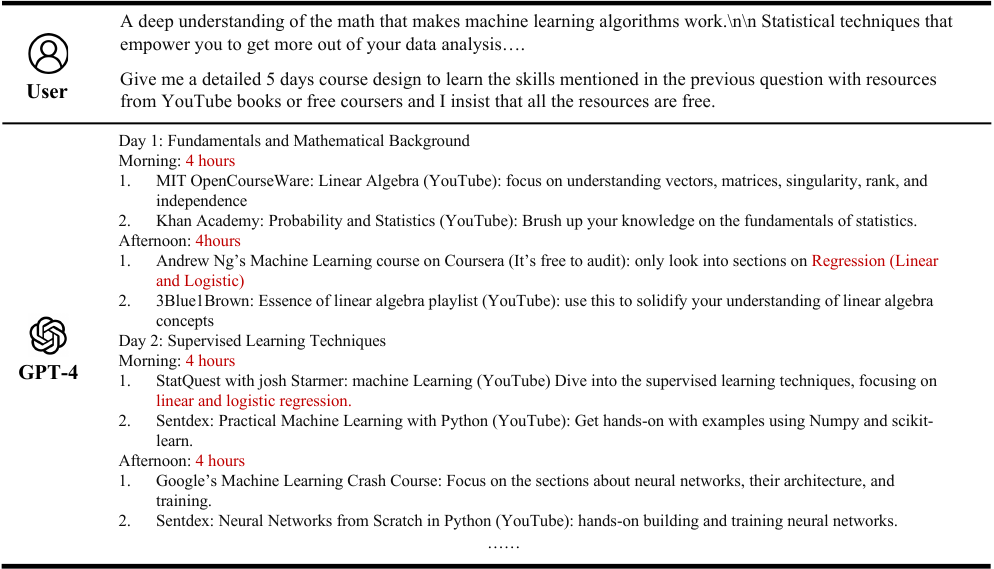}
\caption{An example of course design and the corresponding GPT response.}
\label{fig:course_design}
\vspace*{-3mm}
\end{figure*}

\begin{figure*}[!t]
\centering
\includegraphics[width=0.98\textwidth]{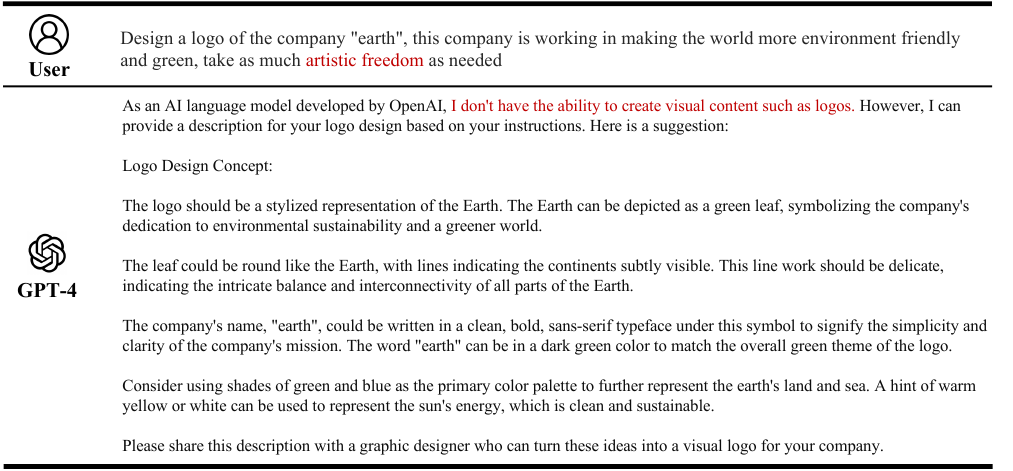}
\caption{An example of logo design and the corresponding GPT response.}
\label{fig:logo_design}
\vspace*{-3mm}
\end{figure*}

In Figure~\ref{fig:course_design} and Figure~\ref{fig:logo_design} we visualize two examples related to ``design''. For course design in Figure~\ref{fig:course_design}, the user requires a 5-day design to learn the required content. From GPT-4 response, we can see that the time scheduling is still not feasible to accommodate the content listed, and the design could be overlapped on certain topics. For logo design in Figure~\ref{fig:logo_design}, GPT-4 does not hold the multi-modal ability to deal with this query, and in the conceptual design description, we can see that the design may not fulfill aesthetic beauty requirements and lacks creativity.

\begin{figure*}[!t]
\centering
\includegraphics[width=0.98\textwidth]{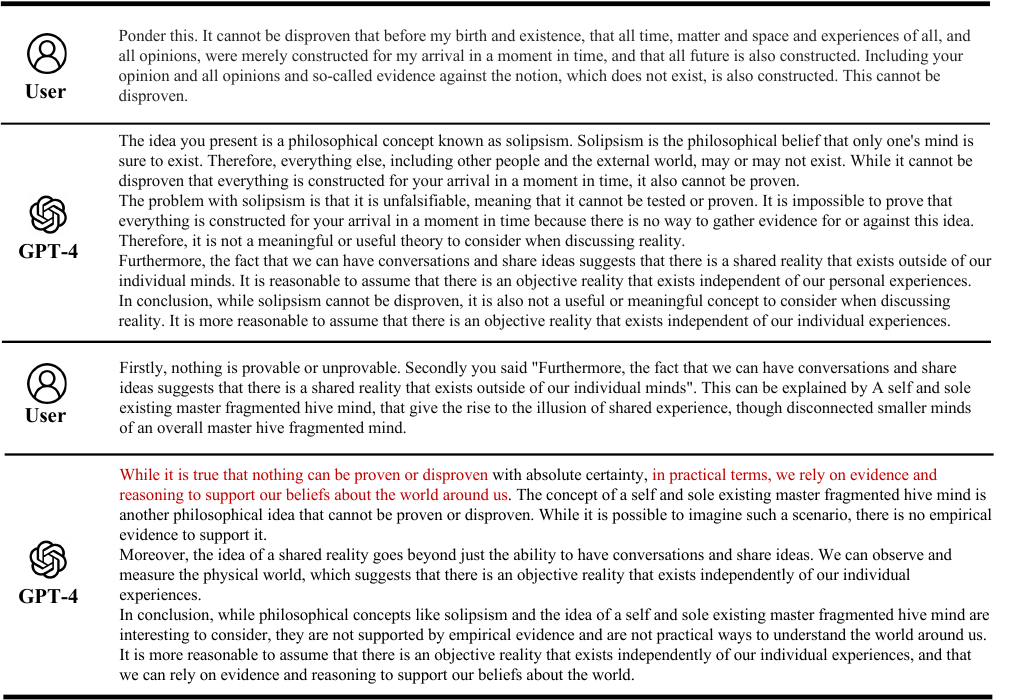}
\caption{An example of philosophical discussion with GPT-4.}
\label{fig:philosophical_discussion}
\vspace*{-3mm}
\end{figure*}

In Figure~\ref{fig:philosophical_discussion} we present another example of ``discussion'', where the user is initiating a philosophical discussion with GPT. We can see that of the most time, GPT is quoting from existing philosophical viewpoints, and sometimes even demonstrates contradictions in the responses due to the conflict between user input and the retrieved opinions. 

\begin{figure*}[!t]
\centering
\includegraphics[width=0.98\textwidth]{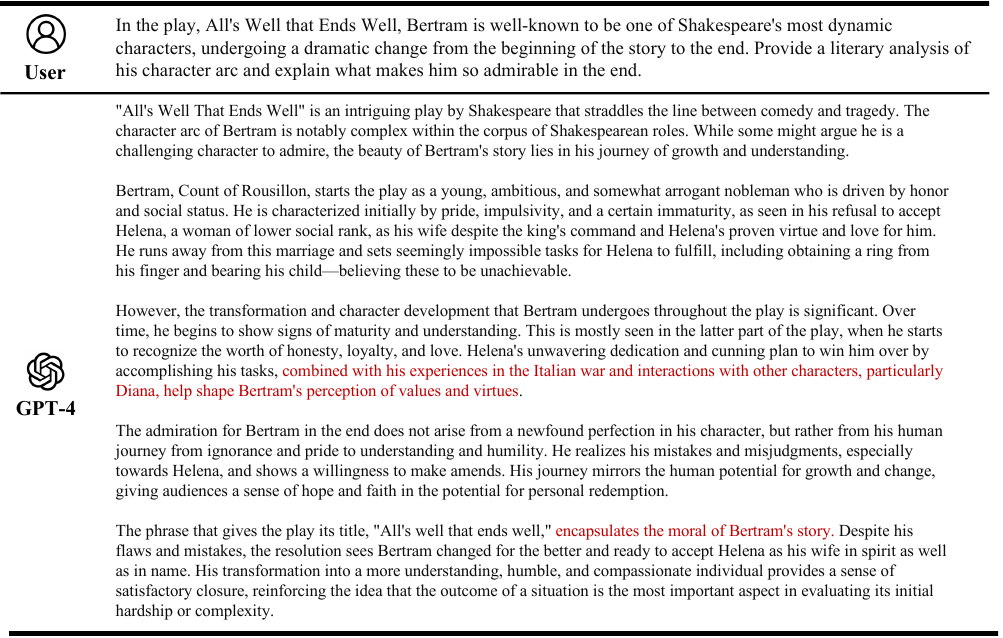}
\caption{An example of literary analysis of a character with GPT-4.}
\label{fig:literary_analysis}
\vspace*{-3mm}
\end{figure*}

Figure~\ref{fig:literary_analysis} shows an example of ``analysis'' concerning a character in the play written by Shakespeare. This is a very open and controversial topic for analysis. While in GPT's response, we observe hallucinations, especially towards the change of personality of Bertram, and the misinformation of plots when the character is in Italy.

\end{document}